\documentclass[letterpaper]{article} 
\usepackage{aaai24}  
\usepackage{times}  
\usepackage{helvet}  
\usepackage{courier}  
\usepackage[hyphens]{url}  
\usepackage{graphicx} 
\usepackage{natbib}  
\usepackage{caption} 
\usepackage{epsfig}
\usepackage{amsmath}
\usepackage{amssymb}

\usepackage{multirow}
\usepackage{tikz}
\usepackage{comment}
\usepackage{diagbox}
\usepackage{booktabs}
\usepackage{threeparttable}

\usepackage{algorithm}
\usepackage{algorithmic}

\usepackage[capitalize]{cleveref}
\Crefname{table}{Table}{Tables}
\crefname{table}{Tab.}{Tabs.}
\Crefname{figure}{Figure}{Figures}
\crefname{figure}{Fig.}{Figs.}
\Crefname{equation}{Equation}{Equation}
\crefname{equation}{Eq.}{Eq.}

\usepackage{newfloat}
\usepackage{listings}
\DeclareCaptionStyle{ruled}{labelfont=normalfont,labelsep=colon,strut=off} 
\floatstyle{ruled}
\newfloat{listing}{tb}{lst}{}
\floatname{listing}{Listing}
\title{Self-distillation Regularized Connectionist Temporal Classification Loss for\\ Text Recognition: A Simple Yet Effective Approach}
\author{
Ziyin Zhang\equalcontrib, Ning Lu\equalcontrib, Minghui Liao, Yongshuai Huang, Cheng Li, Min Wang, Wei Peng
}
\affiliations{
    Huawei Technologies Co., Ltd \\
Shenzhen, China \\
\tt\small \{zhangziyin1, luning12, liaominghui1, huangyongshuai1, \\
\tt\small licheng81, wangmin5, peng.wei1\}@huawei.com
}

\usepackage{bibentry}

\begin{document}

\maketitle

\begin{abstract}
Text recognition methods are gaining rapid development. Some advanced techniques, e.g., powerful modules, language models, and un- and semi-supervised learning schemes, consecutively push the performance on public benchmarks forward. However, the problem of how to better optimize a text recognition model from the perspective of loss functions is largely overlooked. CTC-based methods, widely used in practice due to their good balance between performance and inference speed, still grapple with accuracy degradation. This is because CTC loss emphasizes the optimization of the entire sequence target while neglecting to learn individual characters. We propose a self-distillation scheme for CTC-based model to address this issue. It incorporates a framewise regularization term in CTC loss to emphasize individual supervision, and leverages the maximizing-a-posteriori of latent alignment to solve the inconsistency problem that arises in distillation between CTC-based models. We refer to the regularized CTC loss as \textbf{D}istillation \textbf{C}onnectionist \textbf{T}emporal \textbf{C}lassification (DCTC) loss. DCTC loss is module-free, requiring no extra parameters, longer inference lag, or additional training data or phases. Extensive experiments on public benchmarks demonstrate that DCTC can boost text recognition model accuracy by up to 2.6\%, without any of these drawbacks.






\end{abstract}


\section{Introduction}

\label{sec:intro}

Text Recognition (TR) is an indispensable technology that facilitates intelligent auto-driving~\cite{ZhuLYL18}, revealing precise semantic information~\cite{Chen2021a} for sensitive information auditing, saving labor forces for financial processes, etc. Methods for scene text recognition (STR) are blooming at a breathless pace these years. For example, \cite{SVTR, DBLP:conf/eccv/DaWY22, Lu_2021} focus on designing sophisticated architectures by inventing powerful modules;~\cite{DBLP:conf/cvpr/WangLCGLRB022, YuxinWang2022PETRRT} integrate a language model into a text recognition model to enable explicit language modeling;~\cite{DBLP:conf/wacv/PatelAQ23, DiG} learn better sequential features with an un-supervised or semi-supervised learning scheme by leveraging a large amount of label-free or partial labeled data; However, the problem that how to better optimize a text recognition model from a perspective of loss functions is out of in the cold. It is also worth lots of effort since the dedicated designed loss function may be free of extra parameters, extra inference latency, extra training data, or extra training phases.

\begin{figure}[t]
     \centering
     \includegraphics[width=\columnwidth]{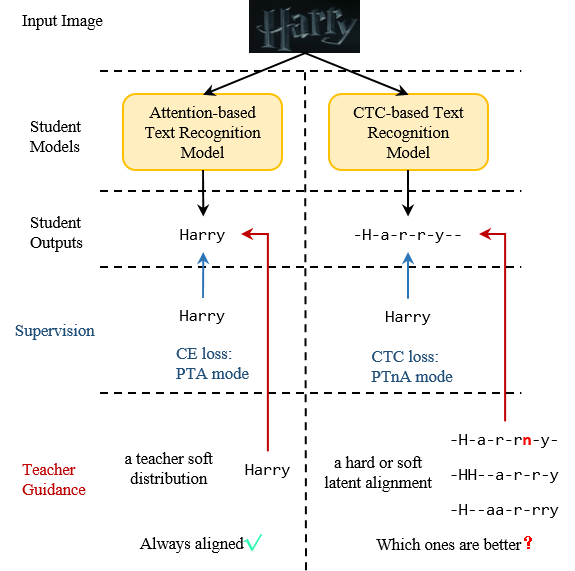}
        \caption{An illustraion of optimization and distillation on CTC- and attention-based models. Also shows the alignment inconsistency problem}
        \label{fig:arch}
\end{figure}

Recent text recognition methods are often supervised by two loss functions, the Connectionist Temporal Classification~(CTC) loss and the Cross-Entropy~(CE) loss, which correspond to CTC-based and attention-based models, respectively. As illustrated in \cref{fig:arch}, CTC loss and CE loss optimize models in a Prediction-Target non-Aligned~(PTnA) and Prediction-Target Aligned~(PTA) mode respectively. Although much recent research empirically shows that CE-based models, which run in the PTA mode, can outperform CTC-based models~\cite{DBLP:conf/icdar/CongHHG19, DBLP:conf/cvpr/ShiWLYB16, JeonghunBaek2019WhatIW}, CTC-based models have three non-negligible advantages: 1) The CTC decoder is more robust to varying input sequence lengths than the attention-based decoder~\cite{DBLP:conf/icdar/CongHHG19, benchmark_cn}; 2) Compared to attention-based models, CTC-based models, getting rid of auto-regression, decode each time-step simultaneously, which can achieve better inference efficiency~\cite{DBLP:journals/ijcv/LongHY21, benchmark_cn}; 3) Due to its concise model design~\cite{https://doi.org/10.48550/arxiv.2206.03001, 10.1145/3474085.3478328}, CRNN~\cite{CRNN}, the classical CTC-based TR model, is still a mainstream industrial model. These practical advantages attract our research focus back to the CTC loss function, motivating this paper.


CTC loss models the negative log total probability of all feasible paths that can be collapsed into the label sequence. However, some of the paths are more plausible. These paths are certain particular alignments of all positions along the sequence. Once discovered and additionally trained with such alignments, the model should be benefited from that. The process of discovering and training those more plausible alignments is known as Knowledge Distillation (KD)~\cite{GeoffreyEHinton2015DistillingTK}. The more plausible alignments are also a form of ``dark knowledge"~\cite{GeoffreyEHinton2015DistillingTK} in the context of KD. Nevertheless, a common issue when applying KD to CTC-based models is ``alignment inconsistency"~\cite{HaisongDing2020ImprovingKD}. This issue occurs when the features or outputs of the teacher model are found to be inaccurate or inconsistent. This inconsistency can arise due to the limitations of the teacher alignment, which cannot guarantee full correctness or consistency during training or across multiple teacher models. As a result, this can negatively impact the performance of the distillation process.

The key to success in distillation on CTC models is to find the proper alignments, i.e., the latent alignments. Previous works~\cite{HaisongDing2020ImprovingKD, MingkunHuang2018KnowledgeDF} are module-dependent. They estimate the latent alignment directly from other teacher models' outputs or intermediate features. To obtain more accurate latent alignments, these methods often require complex and well-trained teacher models. To further stabilize the estimate, some~\cite{DBLP:conf/emnlp/KimR16,DBLP:conf/interspeech/Ding0H19} use specifically designed heuristic mechanism to adjust the original one or use an ensemble of a group of raw estimates. Some methods~\cite{DBLP:conf/slt/KurataA18,DBLP:conf/interspeech/Ding0H19} use an ensemble of teachers to improve guidance accuracy. However, 1)they used extra complex teacher models, which increases computing resource demand; 2) they can hardly relieve the intrinsic inaccuracy as a result of directly taking the outputs of the teacher models as the latent alignment; and 3) they incurred distillation instability when using ensembles of teachers because of inconsistent peak positions~\cite{DBLP:conf/slt/KurataA18}, causing unstable collapsed latent alignments.

We propose \textbf{D}istillation \textbf{CTC} loss (DCTC), a frame-wise, self-distillation scheme for the CTC-based models. By modeling latent alignment distribution as maximizing the posterior probability given the ground truth and model outputs, we derive a simple, effective, and module-free method to generate high-quality estimated latent alignment at each training iteration. This method is closed-formed and does not require any additional module to perform. In summary, our contributions are as follows:


\begin{enumerate}
\item  We propose a self-distillation scheme, DCTC, to conduct frame-wise regularization for CTC-based models. It can directly apply to existing CTC-based text recognition model without introducing extra teacher models, training phases, or training data.

\item To our knowledge, it is the first work that uses MAP to perform latent alignment estimate. Our method well addresses the alignment inconsistency problem by generating high-quality estimated latent alignment most of the training time, which is supported by our quantitative analysis.

\item Exhaustive experiments over models and CTC loss variants demonstrate that our proposed DCTC loss effectively boost the performance of various text recognition models on both English and Chinese text recognition benchmarks.

\end{enumerate}



\section{Related works}

\subsection{Text recognition}\label{RW:TR}
Text Recognition is vital in the Optical Character Recognition (OCR) area. In the deep-learning era, how to design powerful modules attracts lots of interest. Shi et al.~\cite{CRNN} proposed a segmentation-free method, CRNN, which models sequential relationships between frames and employs CTC loss~\cite{10.1145/1143844.1143891}, adaptively aligning features to targets to train a neural network. This method gained huge success and opened a new era for STR. ~\cite{SVTR} used ViT~\cite{DBLP:conf/iclr/DosovitskiyB0WZ21} to develop a single powerful visual model for recognition. It also employs CTC loss to align targets. \cite{Lu_2021,fang2021read,yu2020accurate,DBLP:conf/iccv/BhuniaSKGCS21} formulated text recognition problem as a translation task that translates a cropped image into a string, using an encoder-decoder framework, along with an attention mechanism~\cite{JeonghunBaek2019WhatIW}. Recently, thanks to Self-Attention~\cite{AshishVaswani2017AttentionIA}, \cite{Lu_2021, MinghaoLi2021TrOCRTO} proposed transformer-based STR models to solve the attention drift problem~\cite{ZhanzhanCheng2017FocusingAT}. 
Besides, Liao et al.~\cite{CA-FCN} proposed to segment and recognize text from two-dimensional perspective.
Another active direction is to lay their hope in a language model. \cite{DBLP:conf/cvpr/QiaoZYZ020} claimed that the encoder-decoder framework only focuses on the local visual feature while ignoring global semantic information. So they used a pre-trained language model to guide the decoding process to improve the model's performance. \cite{fang2021read} integrated a language model into a vision-based recognition model to enhance its feature representative ability, which iteratively refines the model's prediction. A more advanced work by Bautista et al.~\cite{DBLP:conf/eccv/BautistaA22} used permutation language modeling to refine recognition results. To leverage large unlabelled data, \cite{DBLP:conf/cvpr/AberdamLTASMMP21} proposed a contrastive pre-training learning scheme to boost performance. Recently, \cite{SIGA, CCD, DiG} used self-supervision framework to refine visual and language features at a fine-granularity level to improve recognition accuracy.

\subsection{CTC-related text recognition methods}\label{RW:CTCloss}

Many endeavors have been devoted to improving CTC-based text recognition models. ~\cite{FocalCTC} proposed FocalCTC to aim at the imbalance problem of Chinese words, introducing loss~\cite{FocalLoss} into CTC loss to modulate the importance of hard and easy word examples. Naturally, CTC loss is not designed for 2D spatial prediction. Xie et al.~\cite{DBLP:conf/cvpr/XieHZJLX19} proposed an easy-to-apply aggregated cross entropy (ACE) loss to better solve 2D prediction problems with fast and lightweight implementation.  \cite{DBLP:journals/corr/abs-1907-09705} extended vanilla CTC as 2D-CTC to adapt to 2D text images by modeling a spatial decoding strategy. To encourage cohesive features, Center loss~\cite{YandongWen2016ADF} is introduced to CTC loss as Center-CTC loss~\cite{ppocrv2}. ~\cite{EMCTC} provided an expectation-maximum view of CTC loss and a novel voting algorithm to improve decoding performance. Based on maximum entropy regularization~\cite{ETJaynes1957InformationTA}, ~\cite{EnCTC} proposed EnCTC to address peaky distribution problem~\cite{10.1145/1143844.1143891}. VarCTC~\cite{VarCTC} is also proposed to relieve the problem. Tanaka et al. ~\cite{FDSCTC} used the framework of virtual adversarial training~\cite{TakeruMiyato2017VirtualAT} to develop a fast regularization algorithm FDS on CTC loss by smoothing posterior distributions around training data points.

\subsection{Knowledge distillation on text recognition or CTC-based models}\label{RW:KD-CTC}

There are a lot of works attempted to apply KD on TR models or CTC-based models. ~\cite{DBLP:conf/iccv/BhuniaSCS21} creatively employed a knowledge distillation loss to train a unified model for scene and handwritten TR tasks. However, this method needs two additional teacher models, leading to a complicated training procedure. ~\cite{DBLP:conf/icassp/TakashimaLK18} investigated frame- and sequence-level KD on CTC-based acoustic models. ~\cite{DBLP:conf/emnlp/KimR16} proposed a word-level and a sequence-level distillation method and apply them to neural machine translation task. They used beam search to generate hypotheses from output probabilities and kept a K-best list to approximate the teacher distribution. ~\cite{DBLP:conf/interspeech/Ding0H19} used N-best hypotheses imitation to do frame- and segment-wise distillation from a complex teacher model. ~\cite{DBLP:conf/slt/KurataA18} proposed an alignment-consistent ensemble technique to relieve unstable ensemble alignment problem. ~\cite{DBLP:conf/interspeech/MoriyaOKSTAMSD20} uses self-distillation KD on the CTC-based ASR system by using a Transformer~\cite{AshishVaswani2017AttentionIA} module to generate latent alignment. Recently, CCD~\cite{CCD} used a self-distillation module to perform character-level distillation. SIGA~\cite{SIGA} used a self-supervised implicit glyph attention module to relief the alignment-drifted issue, which can be also seen as an character-level self-distillation. The aforementioned methods are all module-dependent and need extra teacher models to provide accurate estimated latent alignment. Also, they can hardly give a closed form for the estimated latent alignments.



\section{Methods}

\begin{figure*}[th]
  \centering
   \includegraphics[width=2\columnwidth]{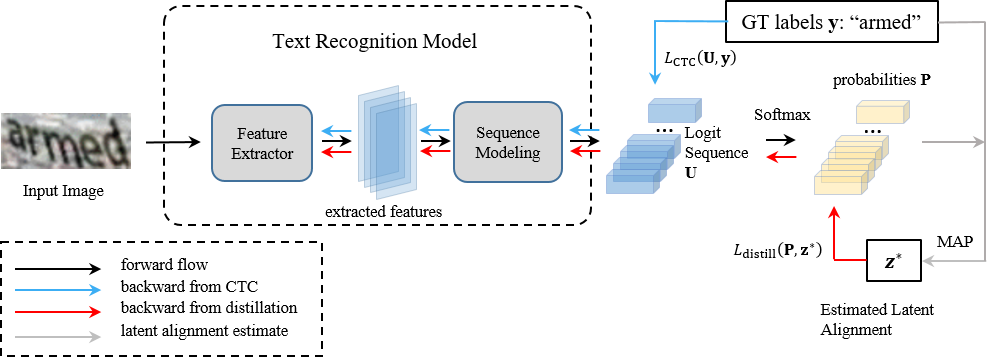}
    \caption{The Architecture of DCTC in Self-distillation Scheme}
    \label{fig:method}
\end{figure*}

A key problem in CTC distillation is alignment inconsistency~\cite{DBLP:conf/slt/KurataA18}. The problem can be described as to find a proper latent alignment $\mathbf{z}\in {V'}^T$, from which the student model can distill and whose length is equal to that of the logits sequence $\mathbf{U}\in \mathbb{R}^{K+1,T}$. $V$ is the character vocabulary, $V'=V\cup\{{\rm blank}\}$ is the augmented vocabulary in the CTC setting, and there are $|V|=K$, $|V'|=K+1$. $L$ is the length of the label sequence, $T$ is the number of time steps, or the length of the logit sequence. It is required that $T>L$ in the CTC setting, causing non-unique alignment, which is the source of alignment inconsistency. We need a way to estimate proper alignments for $\mathbf{U}$ to perform frame-wise KD, which motivates our work.

\subsection{The Distillation Loss Term in CTC Scenario}

Given a sequence of the logits sequence $\mathbf{U}\in \mathbb{R}^{K+1,T}$, the output probability sequence $\mathbf{P}={\rm Softmax}_{V'} (\mathbf{U})$, the ground truth label sequence $\mathbf{y}\in V^L$, and the latent alignment $\mathbf{z}\in {V'}^T$. The true label sequence $\mathbf{y}$ can be regarded as an oracle teacher from which we want the student logits sequence to distill. Originally it was impossible because the lengths of the true label sequence and the logits sequence are different ($L<T$). However, bridging by $\mathbf{z}$ as an agent, the distillation loss term can be formulated as:

\begin{equation}
\label{L_distill}
    \centering
    \mathcal{L}_{\rm distill}(\mathbf{P},\mathbf{z})= \mathcal{L}_{\rm CE}(\mathbf{P},\mathbf{z})=-\sum_{t=1}^T z_t \log\mathbf{P}(z_t,t)
\end{equation}

Distillation CTC is defined as follows:

\begin{equation}
\label{DCTC}
    \centering
    \mathcal{L}_{\rm DCTC}(\mathbf{U},\mathbf{P},\mathbf{y},\mathbf{z})=\mathcal{L}_{\rm CTC}(\mathbf{U},\mathbf{y})+\lambda\mathcal{L}_{\rm distill}(\mathbf{P},\mathbf{z})
\end{equation}

where $\lambda$ is the coefficient controlling the amplitude of distillation.

The question is how to give a proper latent alignment $\mathbf{z}$. In the self-distillation scheme, it can be generated by a layer or an additional MLP head of the student model. Nevertheless, module-dependent methods often yield bad generation quality, which will be shown in experiments. Aiming to solve this problem, instead of using an module-dependent method, we deduce a closed-form estimation of $\mathbf{z}$ via Maximum-A-Posteriori (MAP).

\subsection{Estimation of Latent Alignment $\mathbf{z}$}\label{pseudo_label_generation}

For every $t$ from 1 to $T$, we want to find a certain value for $z_t$, which can most likely be decoded into the given true label sequence $\mathbf{y}$. Denote the best estimation of $\mathbf{z}$ as $\mathbf{z}^*$, then $\mathbf{z}^*$ is given by

\begin{equation}
\label{z_star}
    \centering
    \mathbf{z}^*=\mathop{\arg\min}_{V'} \frac{\mathbf{G}}{\mathbf{P}}
\end{equation}

where $\mathbf{G}$ is the gradient tensor of CTC loss with respect to the logits sequence, that is:

\begin{equation}
\label{G_definition}
    \centering
    \mathbf{G}=\frac{\partial\mathcal{L}_{\rm CTC}(\mathbf{U},\mathbf{y})}{\partial\mathbf{U}}
\end{equation}

\textbf{Derivation of our generation method} Given input (image) $\mathbf{X}$ and its corresponding true label sequence $\mathbf{y}$, at time $t$, the certain $z_t$ that is most likely to decoded into $\mathbf{y}$ can be formulated as a MAP estimate:

\begin{align}
\label{map_def}
    \centering
    \begin{split}
    z_t^*&=\mathop{\arg\max}_{z_t\in V'} p(\mathbf{y}|z_t,\mathbf{X}) \\
&= \mathop{\arg\max}_{z_t\in V'} \frac{p(z_t|\mathbf{y},\mathbf{X})p(\mathbf{y}|\mathbf{X})}{p(z_t|\mathbf{X})} \\
&= \mathop{\arg\max}_{z_t\in V'} \frac{p(z_t|\mathbf{y},\mathbf{X})}{p(z_t|\mathbf{X})}
    \end{split}
\end{align}

In \cref{map_def}, $p(z_t|\mathbf{X})$ is the probability that is directly output by the model. $p(z_t|\mathbf{X},\mathbf{y})$ is the probability that character $z_t\in V'$ appears at time step $t$ when $\mathbf{y}$ and $\mathbf{X}$ are given. We now model $p(z_t|\mathbf{X},\mathbf{y})$ in the CTC setting. Using \cite{10.1145/1143844.1143891}'s notation, let $\alpha(\cdot,\cdot)$ and $\beta(\cdot,\cdot)$ be the forward and the backward table respectively. $\alpha(\cdot,\cdot),~ \beta(\cdot,\cdot)\in\mathbb{R}^{l',T}$, where $l'=2L+1$ is the length of the augmented true label sequence $\mathbf{y}'\in {V'}^{l'}$ used in computing $\alpha$ and $\beta$. Forward table element $\alpha(i,t)$ means the probability that the cumulative paths go through $\mathbf{y}'_i\in V'$ at time step $t$ from the start of $\mathbf{y}'$. Backward table element $\beta(i,t)$ means the probability that the cumulative paths go through $\mathbf{y}'_i$ at time step $t$ from the end of $\mathbf{y}'$. As such, $\alpha(i,t)\beta(i,t)/\mathbf{P}(\mathbf{y}'_i,t)$ is the probability that the total paths go through $\mathbf{y}'_i$ at time step $t$ over the whole time sequence. Denote $S(i,t)=\alpha(i,t)\beta(i,t)/\mathbf{P}(\mathbf{y}'_i,t)$ for simplicity. Then, for a specific class $c$, in the CTC setting, we have:

\begin{equation}
\label{p_z_X_y}
    \centering
    p(z_t=c|\mathbf{X},\mathbf{y})\propto\sum_{i,\mathbf{y}'_i=c}^{l'} S(i,t)
\end{equation}

We need a way to connect \cref{p_z_X_y} to a value that we can easily compute. Observe that the gradients of CTC loss with respect to $\mathbf{U}$ is given by~\cite{10.1145/1143844.1143891}:

\begin{equation}
\label{ctc_grad}
    \centering
    \frac{\partial\mathcal{L}_\mathrm{CTC}}{\partial\mathbf{U}(c,t)}=\mathbf{G}(c,t)=\mathbf{P}(c,t)-\frac{\sum_{i,\mathbf{y}'_i=c}^{l'} S(i,t)}{p(\mathbf{y}|\mathbf{X})}
\end{equation}

So we have:

\begin{equation}
\label{p_z_X_y_2}
    \centering
\begin{split}
p(z_t=c|\mathbf{X},\mathbf{y}) &\propto (\mathbf{P}(c,t)-\mathbf{G}(c,t))p(\mathbf{y}|\mathbf{X}) \\
&\propto \mathbf{P}(c,t)-\mathbf{G}(c,t)
\end{split}
\end{equation}

Note the fact that $\mathbf{P}(c,t)=p(z_t=c|\mathbf{X})$. Now, comparing \cref{map_def} and \cref{p_z_X_y_2}, \cref{map_def} becomes:

\begin{equation}
\label{map_def_2}
    \centering
\begin{split}
    z_t^* & = \mathop{\arg\max}_{c\in V'}\frac{\mathbf{P}(c,t)-\mathbf{G}(c,t)}{\mathbf{P}(c,t)} \\
& = \mathop{\arg\max}_{c\in V'}\left(1-\frac{\mathbf{G}(c,t)}{\mathbf{P}(c,t)} \right) \\
& = \mathop{\arg\min}_{c\in V'}\frac{\mathbf{G}(c,t)}{\mathbf{P}(c,t)}
\end{split}
\end{equation}

The ultimate form \cref{z_star} is simply the vectorized version of \cref{map_def_2}, which is easy to implement. \textbf{We use \cref{z_star} to generate latent alignment in practice.}

There might be a concern that \cref{z_star} seems to have singularities when $\mathbf{P}$ has zeros, which impedes the calculation of $\mathbf{z}^*$. However, it is not the case. There is NO singularity at all. We can prove that $\mathbf{G}(c,t)/\mathbf{P}(c,t)$ is bounded within $[0,1]$ along $\mathbf{P}(c,t)$ changing from 0 to 1. The proof, however, is cumbersome. Readers who are interested in it can refer to supplementary materials.
 
Our proposed estimation method can generate incredibly high-quality latent alignment. We empirically show that in experiments.

\subsection{Summary of DCTC loss}

DCTC loss works in a self-distillation scheme, as such, $\mathbf{z}^*$ is directly estimated from the CTC loss that supervises the student model. No other teacher models participated. So, we substitute \cref{z_star} into \cref{DCTC} and get: 

\begin{equation}
\label{DCTC_final}
    \centering
    \mathcal{L}_{\rm DCTC}(\mathbf{U},\mathbf{P},\mathbf{y},\mathbf{z}^*)=\mathcal{L}_{\rm CTC}(\mathbf{U},\mathbf{y})+\lambda\mathcal{L}_{\rm distill}(\mathbf{P},\mathbf{z}^*)
\end{equation}

We show a pseudo code of DCTC loss in \cref{pc_lmr_ctc} for a clear understanding. Meanwhile, the architecture of our method is shown in \cref{fig:method}.

\begin{algorithm}
	\renewcommand{\algorithmicrequire}{\textbf{Input:}}
	\renewcommand{\algorithmicensure}{\textbf{Output:}}
	\caption{Calculation of DCTC loss in self-distillation scheme}
	\label{pc_lmr_ctc}
	\begin{algorithmic}[1]
	\REQUIRE the input logits $\mathbf{U}$, ground truth label sequence $\mathbf{y}$, weighting factor $\lambda$
        \STATE Calculate probabilities $\mathbf{P}=\mathrm{softmax}_{V'} \mathbf{U}$.
        \STATE Calculate CTC loss $\mathcal{L}_1=\mathcal{L}_{\rm CTC} (\mathbf{U,y})$.
        \STATE Without tracing gradients, copy $\mathbf{U}$ as $\mathbf{U'}$
        \STATE Calculate CTC loss $\mathcal{L}_2=\mathcal{L}_{\rm CTC} (\mathbf{U',y})$.
        \STATE Calculate gradients $\mathbf{G}=\partial\mathcal{L}_2/\partial \mathbf{U'}$
        \STATE Take argmin over vocabulary: $\mathbf{z}^*=\mathop{\arg\min}_{V'} \mathbf{G}/\mathbf{P}$
        \STATE Compute $\mathcal{L}_{\rm DCTC}=\mathcal{L}_1+\lambda\mathcal{L}_{\rm distill}(\mathbf{P},\mathbf{z}^*)$
	\ENSURE  $\mathcal{L}_{\rm DCTC}$
	\end{algorithmic}  
\end{algorithm}

\section{Experiments}

\subsection{Datasets}\label{datasets}
All datasets used in our experiments are publicly available. Our experiments are conducted on English and Chinese scenarios. For English text recognition task, we train all models on two commonly used synthetic scene text recognition datasets: ST~\cite{ST} and MJ~\cite{MJ}. We evaluate all models on six English benchmark datasets: IC13~\cite{IC13}, IC15~\cite{IC15}, SVT~\cite{SVT}, SVTP~\cite{SVTP}, IIIT~\cite{IIIT5k} and CT~\cite{CUTE80}. Each of these six contains 857, 647, 3000, 1811, 645 and 288 test samples, respectively. For Chinese text recognition task, we use the Chinese Benchmark datasets~\cite{benchmark_cn}. It contains four subsets: Scene, Web, Document(Doc) and Handwritten(Hand). Each of these four contains 63646, 14059, 50000 and 18651 test samples, respectively. We train all models on their own training set and evaluate them on their own test set. The license of academically using Handwritten subset, aka SCUT-HCCDoc~\cite{SCUT-HCCDoc}, has been issued by its owner as per our request.

\subsection{Implementation Details}\label{exp_details}

\textbf{Base Models} We choose six base models as the student models. They are CRNN~\cite{CRNN}, TRBA~\cite{TRBA}, SVTR-T, SVTR-S and SVTR-B~\cite{SVTR}. We use them trained with CTC loss as the baseline models and compare them to models trained with DCTC loss in a self-distillation scheme (directly replacing CTC loss with DCTC loss), meaning the teacher is the student itself. All models are implemented with PaddleOCR\footnote{\url{https://github.com/PaddlePaddle/PaddleOCR}}. We implemented DCTC as a CUDA-CPP extension for computation efficiency.

\textbf{Hyperparameters} Hyperparameters for the number of training epochs, batch size, data augmentation strategy, optimizer, learning rate, and decay policy are different as per the base models and follow the base models' own original settings described in their source. The image size used for English task is (h,w)=(32,100), and for Chinese task is (32,256). The distillation coefficient $\lambda$ in $\mathcal{L}_{\rm DCTC}$ is set to 0.025 for English tasks, and 0.01 for Chinese tasks. All experiments are conducted on Nvidia Tesla V100 GPUs.

\textbf{Metrics and Evaluation Protocols} We use accuracy to evaluate all models' performance. \textbf{Accuracy (ACC)} is the ratio of the number of totally correct predictions over the number of test samples. Certain protocols applied when evaluating. For English tasks, only numbers and letters (case-insensitive) are evaluated. For Chinese tasks, we follow \cite{benchmark_cn}'s conventions: 1) convert full-width characters to half-width characters; 2) convert traditional Chinese characters to simplified Chinese characters; 3) all letters to lowercase, and 4) discard all spaces.

In addition, we propose a new metric ``\textbf{Alignment Accuracy (AACC)}" to measure the quality of the latent alignment estimate. It is defined as the ACC of the \textbf{decoded} latent alignments and the ground truth labels. The difference from evaluating model performance is that we do not apply any protocol when evaluating AACC. We decode latent alignments in a CTC-greedy way, meaning collapsing repeating characters and removing all blanks.

\begin{table*}[h]
\begin{center}
\caption{Results of Model-wise Comparison. \textbf{Bold} ACCs are the model-wise better results. ACC marked by * means those data are not reported and thus reproduced by us. Results on English benchmarks of the baseline models of CRNN and TRBA are reported by ~\cite{TRBA}. Results on Chinese benchmarks of the baseline model of CRNN are reported by ~\cite{benchmark_cn}. Results of the baseline model of SVTR series are reported by ~\cite{SVTR}.}
\label{table:model_wise_comparison}
\scalebox{0.82}{
\begin{tabular}{|c|c|c|ccccccl|ccccl|}
\hline
\multirow{2}{*}{\begin{tabular}[c]{@{}c@{}}Base\\ Model\end{tabular}} & \multirow{2}{*}{Methods} & \multirow{2}{*}{Venue} & \multicolumn{7}{c|}{English Benchmarks} & \multicolumn{5}{c|}{Chinese Benchmarks} \\ \cline{4-15} 
 &  &  & IC13 & SVT & IIIT & IC15 & SVTP & \multicolumn{1}{c|}{CT} & \multicolumn{1}{c|}{Avg} & Scene & Web & Doc & \multicolumn{1}{c|}{Hand} & \multicolumn{1}{c|}{Avg} \\ \hline
 \hline
\multirow{2}{*}{CRNN} & CTC & TPAMI'15 & 90.3 & 78.9 & 84.3 & 65.9 & 64.8 & \multicolumn{1}{c|}{61.3} & 77.3 & 54.9 & 56.2 & 97.5 & \multicolumn{1}{c|}{48.0} & 68.7 \\
 & DCTC & - & \textbf{90.7} & \textbf{82.4} & \textbf{88.9} & \textbf{66.1} & \textbf{65.4} & \multicolumn{1}{c|}{\textbf{68.1}} & \textbf{79.9(+2.6)} & \textbf{58.6} & \textbf{57.0} & \textbf{98.0} & \multicolumn{1}{c|}{\textbf{49.7}} & \textbf{70.8(+2.1)} \\ \hline
\multirow{2}{*}{TRBA} & CTC & CVPR'21 & 94.0* & 88.9* & 93.6* & 76.5* & 79.8* & \multicolumn{1}{c|}{84.0*} & 87.3 & 59.6* & 57.8* & 98.2* & \multicolumn{1}{c|}{48.9*} & 71.3 \\
 & DCTC & - & \textbf{94.2} & \textbf{90.4} & \textbf{93.9} & \textbf{78.1} & \textbf{81.3} & \multicolumn{1}{c|}{\textbf{85.8}} & \textbf{88.2(+0.9)} & \textbf{61.1} & \textbf{58.6} & \textbf{99.2} & \multicolumn{1}{c|}{\textbf{49.5}} & \textbf{72.4(+1.1)} \\ \hline
\multirow{2}{*}{SVTR-T} & CTC & IJCAI'22 & 96.3 & 91.6 & 94.4 & 84.1 & 85.4 & \multicolumn{1}{c|}{88.2} & 90.8 & 67.9 & 61.8* & 99.1* & \multicolumn{1}{c|}{47.2*} & 75.3 \\
 & DCTC & - & \textbf{96.4} & \textbf{92.3} & \textbf{95.4} & \textbf{85.3} & \textbf{86.1} & \multicolumn{1}{c|}{\textbf{89.9}} & \textbf{91.7(+0.9)} & \textbf{68.3} & \textbf{63.9} & \textbf{99.2} & \multicolumn{1}{c|}{\textbf{48.1}} & \textbf{75.9(+0.6)} \\ \hline
\multirow{2}{*}{SVTR-S} & CTC & IJCAI'22 & 95.7 & \textbf{93.0} & 95.0 & 84.7 & 87.9 & \multicolumn{1}{c|}{92.0} & 91.6 & 69.0 & 63.9* & 99.2* & \multicolumn{1}{c|}{49.5*} & 76.3 \\
 & DCTC & - & \textbf{96.4} & 92.5 & \textbf{96.2} & \textbf{86.2} & \textbf{88.1} & \multicolumn{1}{c|}{\textbf{92.4}} & \textbf{92.5(+0.9)} & \textbf{70.3} & \textbf{65.8} & \textbf{99.4} & \multicolumn{1}{c|}{\textbf{50.3}} & \textbf{77.3(+1.0)} \\ \hline
\multirow{2}{*}{SVTR-B} & CTC & IJCAI'22 & 97.1 & 91.5 & 96.0 & 85.2 & \textbf{89.9} & \multicolumn{1}{c|}{91.7} & 92.3 & 71.4 & 64.1* & 99.3* & \multicolumn{1}{c|}{50.0*} & 77.5 \\
 & DCTC & - & \textbf{97.1} & \textbf{92.9} & \textbf{96.3} & \textbf{87.2} & 89.6 & \multicolumn{1}{c|}{\textbf{92.1}} & \textbf{93.1(+0.8)} & \textbf{72.2} & \textbf{67.0} & \textbf{99.4} & \multicolumn{1}{c|}{\textbf{50.4}} & \textbf{78.2(+0.7)} \\ \hline
\multirow{2}{*}{SVTR-L} & CTC & IJCAI'22 & 97.2 & 91.7 & 96.3 & 86.6 & 88.4 & \multicolumn{1}{c|}{\textbf{95.1}} & 92.8 & 72.1 & 66.3* & 99.3* & \multicolumn{1}{c|}{50.3*} & 78.1 \\
 & DCTC & - & \textbf{97.4} & \textbf{93.7} & \textbf{96.9} & \textbf{87.3} & \textbf{88.5} & \multicolumn{1}{c|}{92.3} & \textbf{93.3(+0.5)} & \textbf{73.9} & \textbf{68.5} & \textbf{99.4} & \multicolumn{1}{c|}{\textbf{51.0}} & \textbf{79.2(+1.1)} \\ \hline
\end{tabular}
}
\end{center}
\end{table*}

\subsection{A Model-wise Comparison}\label{model_wise_comparison}

We compare DCTC loss with CTC loss on six models mentioned and collect all results in \cref{table:model_wise_comparison}. Each model is compared to its baseline, which is the one trained by CTC loss. We can clearly see that all models achieve accuracy improvement over almost all benchmark datasets, which profoundly verifies the effectiveness of our method at the model level. CRNN, the most classical, representative, and widely-used industrial CTC-based text recognition model, obtains a 2.6\% average accuracy increment on English and 2.1\% on Chinese benchmarks. The advanced CTC-based single-visual-model text recognition method, SVTR series, can also gain accuracy improvement by our method. Up to 0.9\% and 1.1\% average accuracy improvement in English and Chinese are observed when trained with DCTC loss. Besides, as our method does not change the structure of the models, the inference speed remains the same. 

\begin{table*}[h]
\begin{center}
\caption{Results of Loss-wise Comparison. ACC marked by * means those data are not reported and thus reproduced by us; Results of CTC loss and DCTC are the same as in \cref{table:model_wise_comparison}; Results of FocalCTC and EnCTC are all reproduced by us.}
\label{table:loss_wise_comparison}
\scalebox{0.82}{
\begin{tabular}{|c|c|c|ccccccl|ccccl|}
\hline
\multirow{2}{*}{\begin{tabular}[c]{@{}c@{}}Base\\ Model\end{tabular}} & \multirow{2}{*}{Variant} & \multirow{2}{*}{Venue} & \multicolumn{7}{c|}{English Benchmarks} & \multicolumn{5}{c|}{Chinese Benchmarks} \\ \cline{4-15} 
 &  &  & IC13 & SVT & IIIT & IC15 & SVTP & \multicolumn{1}{c|}{CT} & \multicolumn{1}{c|}{Avg} & Scene & Web & Doc & \multicolumn{1}{c|}{Hand} & \multicolumn{1}{c|}{Avg} \\ \hline
 \hline
\multirow{4}{*}{CRNN} & CTC & TPAMI'15 & 90.3 & 78.9 & 84.3 & 65.9 & 64.8 & \multicolumn{1}{c|}{61.3} & 77.3 & 54.9 & 56.2 & 97.5 & \multicolumn{1}{c|}{48.0} & 68.7 \\
 & FocalCTC & Complexity'19 & 89.6 & 80.1 & 81.2 & 65.2 & 63.0 & \multicolumn{1}{c|}{60.2} & 75.6(-1.7) & 54.8 & 56.0 & 97.5 & \multicolumn{1}{c|}{48.3} & 68.7(+0) \\
 & EnCTC & NeurIPS'18 & 90.1 & 81.5 & 85.6 & 64.7 & 62.9 & \multicolumn{1}{c|}{59.0} & 77.1(-0.2) & 49.0 & 50.7 & 97.5 & \multicolumn{1}{c|}{36.6} & 64.2(-4.5) \\
 & DCTC & - & \textbf{90.7} & \textbf{82.4} & \textbf{88.9} & \textbf{66.1} & \textbf{65.4} & \multicolumn{1}{c|}{\textbf{68.1}} & \textbf{79.9(+2.6)} & \textbf{58.6} & \textbf{57.0} & \textbf{98.0} & \multicolumn{1}{c|}{\textbf{49.7}} & \textbf{70.8(+2.1)} \\ \hline
\multirow{4}{*}{SVTR-T} & CTC & IJCAI'22 & 96.3 & 91.6 & 94.4 & 84.1 & 85.4 & \multicolumn{1}{c|}{88.2} & 90.8 & 67.9 & 61.8* & 99.1* & \multicolumn{1}{c|}{47.2*} & 75.3 \\
 & FocalCTC & Complexity'19 & 96.0 & 91.0 & 94.3 & 84.1 & 85.1 & \multicolumn{1}{c|}{87.9} & 90.6(-0.2) & 67.1 & 60.2 & 99.2 & \multicolumn{1}{c|}{46.5} & 74.8(-0.5) \\
 & EnCTC & NeurIPS'18 & 94.9 & 90.8 & 94.5 & 84.3 & 85.4 & \multicolumn{1}{c|}{88.2} & 90.6(-0.2) & 65.9 & 63.7 & 97.9 & \multicolumn{1}{c|}{47.1} & 74.2(-1.1) \\
 & DCTC & - & \textbf{96.4} & \textbf{92.3} & \textbf{95.4} & \textbf{85.3} & \textbf{86.1} & \multicolumn{1}{c|}{\textbf{89.9}} & \textbf{91.7(+0.9)} & \textbf{68.3} & \textbf{63.9} & \textbf{99.2} & \multicolumn{1}{c|}{\textbf{48.1}} & \textbf{75.9(+0.6)} \\ \hline
\end{tabular}
}
\end{center}
\end{table*}

\subsection{A Loss-wise Comparison}\label{loss_wise_comparison}

Our method can be regarded as a variant of CTC loss when working in a self-distillation scheme. Many variants of CTC have been proposed but have yet to be experimented with advanced models or on Chinese benchmarks. In this part, we compare our method with other variants of CTC loss. The chosen variants are FocalCTC\footnote{\url{https://github.com/PaddlePaddle/PaddleOCR}}~\cite{FocalCTC} and EnCTC\footnote{\url{https://github.com/liuhu-bigeye/enctc.crnn}}~\cite{EnCTC} We choose them because they 1) have been peer-reviewed and 2) have public code bases (in footnotes).

We align the hyperparameters with their original settings to make the comparison fair. For FocalCTC loss, $\alpha=1$ and $\gamma=2$; for EnCTC loss, the regularization coefficient $\beta=0.2$. The experiment results are collected in \cref{table:loss_wise_comparison}. We use CRNN and SVTR-T as base models for efficiency. We can see that our method consistently achieves improvements on all benchmarks, further proving our method's effectiveness.

\subsection{Comparison of Latent Alignment Estimate}\label{pseudo_label_estimate_comparison}

\begin{figure*}[h]
\centering
\includegraphics[width=2\columnwidth]{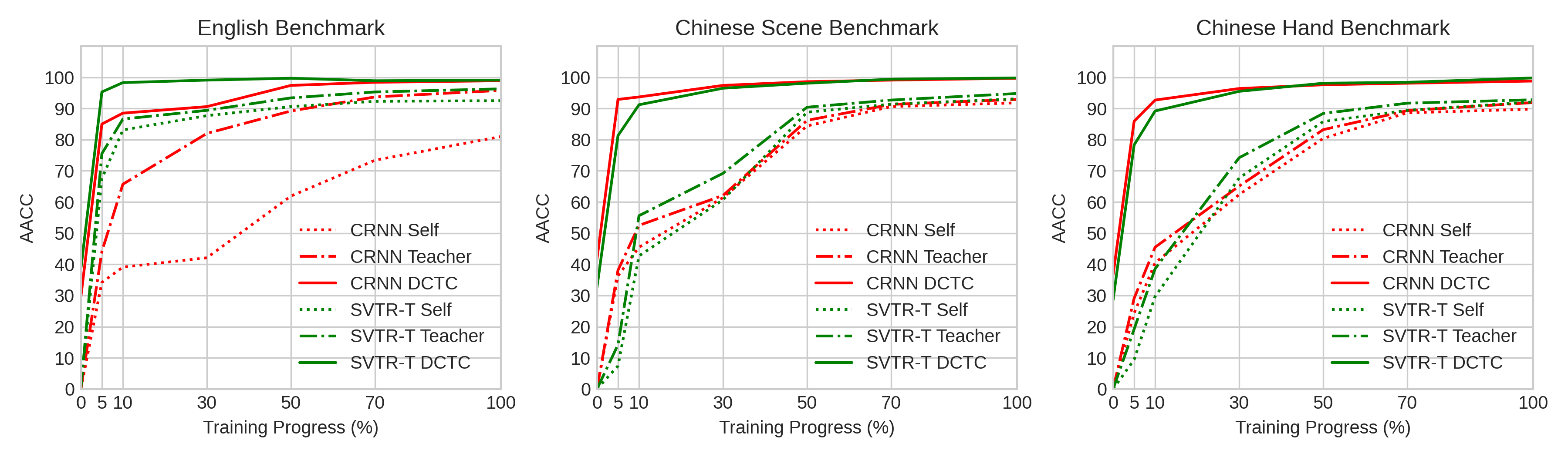}
\caption{Curves of AACC of Estimated Latent Alignment}
\label{fig:aacc}
\end{figure*}

Much previous research on distillation for CTC-based models has been working on finding reasonable estimates of the latent alignment. They used various means to directly to utilize $p(\mathbf{z}|\mathbf{X})$ to estimate the latent alignment. The most naive utilization way is to take the hard prediction of $p(\mathbf{z}|\mathbf{X})$, i.e., $\mathop{\arg\max}_{V'} \mathbf{P}$. In this section, we compare our estimate method with two other sources of estimate: one is to take $\mathop{\arg\max}_{V'} \mathbf{P}$ directly from the model itself, denoted as ``Self". Another is to take $\mathop{\arg\max}_{V'} \mathbf{P}$ from a three-layer Transformer encoder branch additionally added to the model, denoted as ``Teacher". This branch is trained with a CTC loss during the training process. We use CRNN and SVTR-T as the experiment models. We record AACC in training on English, Chinese Scene and, Chinese Hand tasks under different estimate methods, respectively. AACCs are computed by the average over ten consecutive batches at certain progress points in training. Results of AACC are visually shown in \cref{fig:aacc}. We also record the model accuracy under different estimate methods and collect the results in  \cref{table:pseudo_label_method_comparison}.


\begin{table*}[h]
\begin{center}
\caption{Results of latent alignment method estimate method comparison. ACC marked by * means those data are not reported and thus reproduced by us. Recall that ``Self" uses $\mathop{\arg\max}_V \mathbf{P}$ from the model itself, ``Teacher" uses $\mathop{\arg\max}_V \mathbf{P}$ from the added Transformer encoder branch, and ``DCTC" uses $\mathop{\arg\min}_V \mathbf{G}/\mathbf{P}$ from the model itself.}
\label{table:pseudo_label_method_comparison}
\scalebox{0.87}{
\begin{tabular}{|c|c|c|ccccccl|ccccl|}
\hline
\multirow{2}{*}{\begin{tabular}[c]{@{}c@{}}Base\\ Model\end{tabular}} & \multirow{2}{*}{\begin{tabular}[c]{@{}c@{}}Estimate\\ Method\end{tabular}} & \multirow{2}{*}{\begin{tabular}[c]{@{}c@{}}Extra\\ Module\end{tabular}} & \multicolumn{7}{c|}{English Benchmarks} & \multicolumn{5}{c|}{Chinese Benchmarks} \\ \cline{4-15} 
 &  &  & IC13 & SVT & IIIT & IC15 & SVTP & \multicolumn{1}{c|}{CT} & \multicolumn{1}{c|}{Avg} & Scene & Web & Doc & \multicolumn{1}{c|}{Hand} & \multicolumn{1}{c|}{Avg} \\ \hline
 \hline
\multirow{4}{*}{CRNN} & - & - & 90.3 & 78.9 & 84.3 & 65.9 & 64.8 & \multicolumn{1}{c|}{61.3} & 77.3 & 54.9 & 56.2 & 97.5 & \multicolumn{1}{c|}{48.0} & 68.7 \\
 & Self & N & 88.6 & 75.1 & 82.4 & 63.8 & 60.5 & \multicolumn{1}{c|}{59.4} & 75.0(-2.3) & 46.1 & 50.7 & 92.4 & \multicolumn{1}{c|}{40.2} & 61.6(-7.1) \\
 & Teacher & Y & 90.2 & 81.0 & 88.8 & 64.5 & 63.7 & \multicolumn{1}{c|}{65.3} & 79.0(+1.7) & 48.9 & 52.3 & 94.9 & \multicolumn{1}{c|}{42.1} & 64.1(-4.6) \\
 & DCTC & N & \textbf{90.7} & \textbf{82.4} & \textbf{88.9} & \textbf{66.1} & \textbf{65.4} & \multicolumn{1}{c|}{\textbf{68.1}} & \textbf{79.9(+2.6)} & \textbf{58.6} & \textbf{57.0} & \textbf{98.0} & \multicolumn{1}{c|}{\textbf{49.7}} & \textbf{70.8(+2.1)} \\ \hline
\multirow{4}{*}{SVTR-T} & - & - & 96.3 & 91.6 & 94.4 & 84.1 & 85.4 & \multicolumn{1}{c|}{88.2} & 90.8 & 67.9 & 61.8* & 99.1* & \multicolumn{1}{c|}{47.2*} & 75.3 \\
 & Self & N & 95.0 & 90.3 & 93.2 & 84.8 & 85.1 & \multicolumn{1}{c|}{86.5} & 90.1(-0.7) & 67.3 & 60.0 & 99.1 & \multicolumn{1}{c|}{46.6} & 74.8(-0.5) \\
 & Teacher & Y & 95.9 & 91.1 & 94.0 & 83.4 & 85.7 & \multicolumn{1}{c|}{86.4} & 90.3(-0.5) & 66.4 & 61.1 & 99.0 & \multicolumn{1}{c|}{46.5} & 74.5(-0.8) \\
 & DCTC & N & \textbf{96.4} & \textbf{92.3} & \textbf{95.4} & \textbf{85.3} & \textbf{86.1} & \multicolumn{1}{c|}{\textbf{89.9}} & \textbf{91.7(+0.9)} & \textbf{68.3} & \textbf{63.9} & \textbf{99.2} & \multicolumn{1}{c|}{\textbf{48.1}} & \textbf{75.9(+0.6)} \\ \hline
\end{tabular}
}
\end{center}
\end{table*}

We can see from \cref{fig:aacc} that our method (DCTC) can yield high-quality latent alignment (High AACC) even at the beginning of training. ``Teacher" takes second place, while ``Self" only generates moderate estimates after a period of training. Besides, our estimate method gets quickly saturated to nearly 100\% AACC. In contrast, ``Teacher" and ``Self" ling for a long time at a low-to-middle level AACC and can hardly get close to 100\%, not to mention that ``Teacher" used an extra Transformer encoder. \cref{table:pseudo_label_method_comparison} shows that both ``Teacher" and ``Self" cause accuracy degradation on almost all benchmarks, suggesting that simply taking the output of distilled module as the latent alignments is harmful in the CTC setting, no matter whether using a teacher. The only exception is that ``Teacher" boosts CRNN on English benchmarks. We explain that the added Transformer branch fundamentally increases the model capability of CRNN, overcoming the harm from the low-quality estimate of the ``Teacher" method. In conclusion, our method can draw high-quality distillation dark knowledge during the whole training time. This phenomenon explains why DCTC loss can still benefit the student model even under a self-distillation scheme, where no extra teacher model participates.


\subsection{Visual Show of the Effectiveness of Distillation Supervision}

\begin{figure}[htb]
  \centering
  \includegraphics[width=0.75\columnwidth]{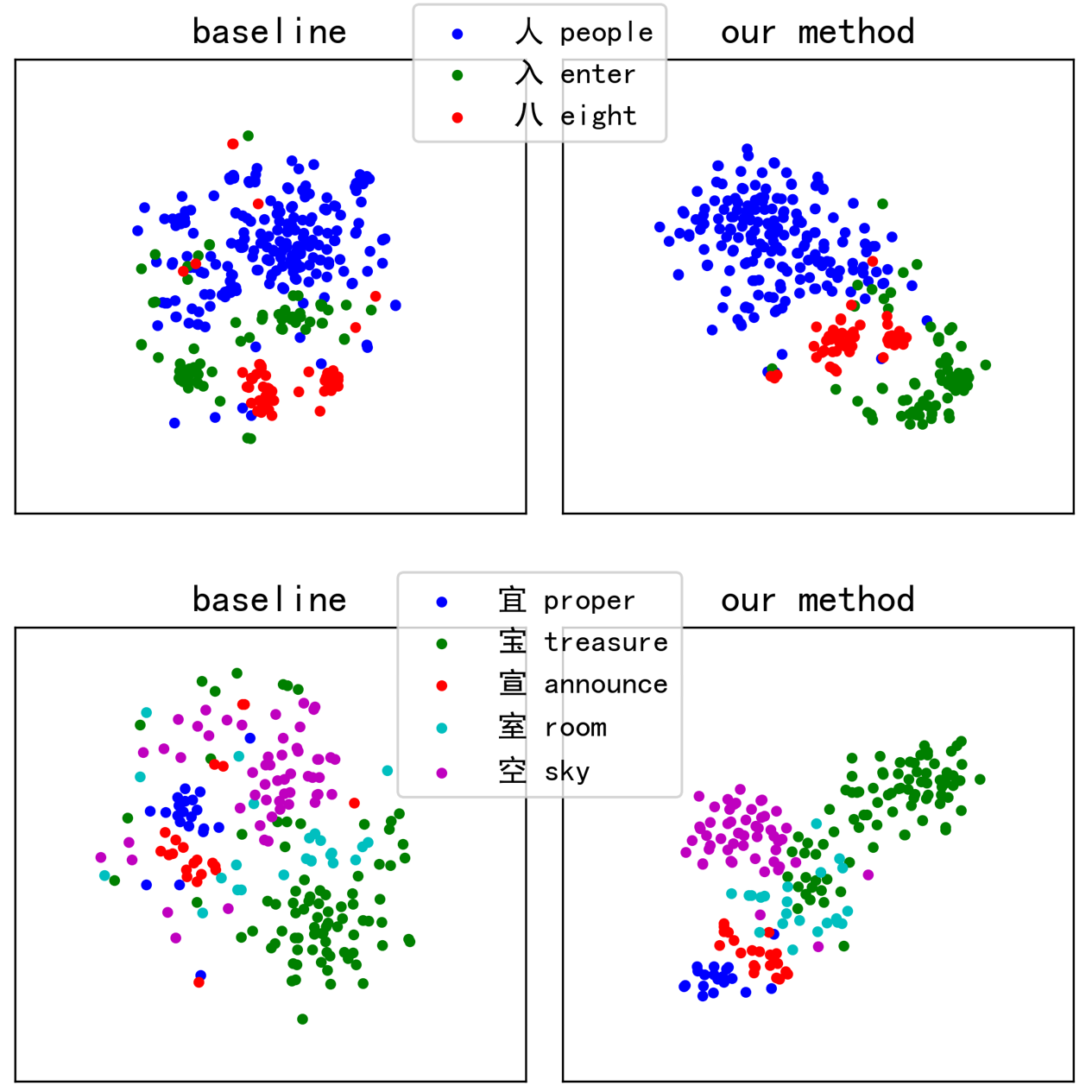}
    \caption{Feature visualization. Each row represents a hard sample cluster}
    \label{fig:feat_proj}
\end{figure}

\cref{L_distill} suggests that DCTC adds frame-wise and character-level supervision to original CTC supervision. Unlike CTC, who more emphasizes sequence supervision, this distillation supervision will make the character features more discriminative, which contributes to the overall performance improvement. We select several hard example clusters from test sets and fetch their features from an SVTR-T model trained with DCTC loss. A hard example cluster is a group of characters more prone to be wrongly recognized as each other. We make a feature visualization study with t-SNE~\cite{Maaten2008VisualizingDU}. \cref{fig:feat_proj} illustrates two hard example clusters of feature projections. Different characters are marked with different colors. Our method drive the model to extract more discriminative features which are more cohesive than those extracted by the baselines. For more clusters, please refer to the supplementary materials.

\section{Conclusion}

In this paper, we base on a self-distillation framework through MAP estimate to formulate DCTC, as a variant of CTC loss. The way we estimate the latent alignments can distill high-quality dark knowledge from the student model itself and well address the alignment inconsistency problem, which is supported by our quantitative analysis. Our proposed DCTC loss is concise yet quite effective. It boasts various text recognition models' performance on both English and Chinese benchmarks. Furthermore, visual analysis shows that DCTC loss can yield more cohesive features, which explains performance improvement. Besides, our method barely incurs additional computational complexity, training data, and training phase. 

\bibliography{aaai24}

\end{document}